\pdfoutput=1

\documentclass[11pt]{article}

\usepackage{acl}

\usepackage{times}
\usepackage{latexsym}

\usepackage[T1]{fontenc}

\usepackage[utf8]{inputenc}

\usepackage{microtype}

\usepackage{inconsolata}

\usepackage{pgf-pie}
\usepackage{pgfplots}

\usepackage{multirow}
\usepackage{arydshln}
\usepackage{graphicx}
\usepackage{subcaption}
\usepackage{hyperref}

\title{StackOverflowVQA: Stack Overflow Visual Question Answering Dataset}

\author{Motahhare Mirzaei, Mohammad Javad Pirhadi \and Sauleh Eetemadi \\
  Iran University of Science and Technology at Tehran, Iran \\
  \texttt{\{m\_mirzaei96, mohammad\_pirhadi\}@comp.iust.ac.ir},
  \texttt{sauleh@iust.ac.ir} \\}

\begin{document}
\maketitle
\begin{abstract}
    In recent years, people have increasingly used AI to help them with their problems by asking questions on different topics. One of these topics can be software-related and programming questions. In this work, we focus on the questions which need the understanding of images in addition to the question itself. We introduce the StackOverflowVQA dataset, which includes questions from StackOverflow that have one or more accompanying images. This is the first VQA dataset that focuses on software-related questions and contains multiple human-generated full-sentence answers. Additionally, we provide a baseline for answering the questions with respect to images in the introduced dataset using the GIT model. All versions of the dataset are available at \href{https://huggingface.co/mirzaei2114}{Hugging Face}.
\end{abstract}

\section{Introduction}

Visual Question Answering \citep{DBLP:journals/corr/AntolALMBZP15} (VQA) aims to answer a question based on an image. Multi-modal generative models and in particular VQA has received a lot of attention from the research community in recent years and has improved significantly \citep{chen2023vqa, li2022mplug, sammani2023uninlx, DBLP:journals/corr/abs-2111-02358}. On the other hand, assisting programmers has also received a lot of attention (e.g., \citealp{friedman2021introducing} introduced GitHub Copilot). Using VQA models, a computer assistant can use the screenshot as the context for a question and better help users.  We introduce the StackOverflowVQA dataset which contains the questions from StackOverflow which has at least one image in their question and one answer. Although single or multi-word answers can be useful in some scenarios, for a computer assistant full sentences or even multiple sentences are often required. However, to the best of our knowledge, datasets with full sentence answers for VQA \citep{shin2016color,sheikhigenerate} have only been artificially synthesized using single or multi word answer datasets such as VQA 2.0 \citep{balanced_vqa_v2}.  StackOverflowVQA is the first large scale human generated full sentence VQA dataset. Training and evaluation of generative tasks such as Machine Translation, Question Answering and VQA is challenging since the same meaning can be expressed in various ways but only a single answer is often available  for each question. Evaluation metrics such as BLEU are also much more accurate when multiple answers are provided. StackOverflowVQA is the first VQA dataset that provides multiple full sentence answers for most questions. Hence this dataset is the best dataset available for training and evaluation of a visually aware computer assistant. While the main contribution of this work is construction of the dataset, we use GIT  \citep{wang2022git} as a state of the art multi-modal generative model which can be used for image/video captioning/question-answering or even image classification. We fine-tune this model on our proposed dataset as a baseline and report the results.

\begin{table*}
\centering
\begin{tabular}{ccccccccc}
\hline
\textbf{Data} & \textbf{B-1} & \textbf{B-2} & \textbf{B-3} & \textbf{B-4} & \textbf{R-1} & \textbf{R-2} & \textbf{R-L} & \textbf{R-LSUM}\\
\hline
StackOverflowVQA-filtered-small & 33 & 18 & 11 & 6.8 & 31 & 5.8 & 23 & 28 \\
\hline
\end{tabular}
\caption{All the results are calculated using the test set of the StackOverflowVQA-filtered-small dataset. (B=BLEU, R=ROUGE)}
\label{tab:results}
\end{table*}

\begin{table}
\centering
\begin{tabular}{lrrr}
\hline
\textbf{Indicator} & \textbf{All} & \textbf{F} & \textbf{FS} \\
\hline
\#Q & 451,394 & 204,041 & 20,458 \\
\cdashline{1-4}
\#A & 799,270 & 204,041 & 20,458 \\
\cdashline{1-4}
\#QwAA & 287,050 & 204,041 & 20,458 \\
\cdashline{1-4}
\#WpQ-min & 8 & 3 & 3 \\
\#WpQ-max & 26,948 & 10,380 & 425 \\
\#WpQ-mean & 299.5 & 267.0 & 122.1 \\
\#WpQ-std & 374.4 & 341.9 & 69.7 \\
\cdashline{1-4}
\#WpA-min & 0 & 0 & 0 \\
\#WpA-max & 8,389 & 8,377 & 425 \\
\#WpA-mean & 151.3 & 167.2 & 87.7 \\
\#WpA-std & 208.3 & 227.0 & 62.7 \\
\cdashline{1-4}
\#ApQ-min & 1 & 1 & 1 \\
\#ApQ-max & 132 & 1 & 1 \\
\#ApQ-mean & 1.8 & 1 & 1 \\
\#ApQ-std & 1.5 & 0 & 0 \\
\cdashline{1-4}
\#PApQ-min & 0 & 1 & 1 \\
\#PApQ-max & 127 & 1 & 1 \\
\#PApQ-mean & 1.1 & 1 & 1 \\
\#PApQ-std & 1.3 & 0 & 0 \\
\cdashline{1-4}
\#IpQ-min & 1 & 1 & 1 \\
\#IpQ-max & 33 & 1 & 1 \\
\#IpQ-mean & 1.4 & 1 & 1 \\
\#IpQ-std & 0.8 & 0 & 0 \\
\hline
\end{tabular}
\caption{Datasets' statistics (F=Filtered, FS=Filtered-Small, Q=Question, A=Answer, QwAA=Questions with Accepted Answer, WpQ=Word per Question, WpA=Word per Answer, ApQ=Answers per Question, PApQ=Positive-scored Answers per Question, IpQ=Images per Question)}
\label{tab:statistics}
\end{table}

\section{Dataset}

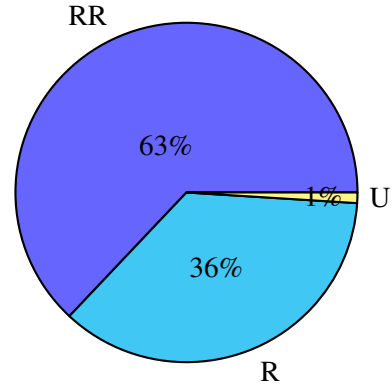
\begin{figure}[t]
\begin{tikzpicture}
  \pie[scale=0.75]{63/RR, 36/R, 1/U}
\end{tikzpicture}
\centering
\caption{Distribution of images with respect to relatedness and requirement. (RR=Related and required, R=Related but not required, U=Unrelated)}
\label{fig:qc}
\end{figure}

\begin{figure*}
\centering
\begin{subfigure}{.45\textwidth}
    \centering
    \includegraphics[width=0.95\linewidth]{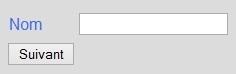}
    \caption{Q: I want to have a box in HTML such as this one: <IMAGE> [rest of question]}
    \label{fig:example_a}
\end{subfigure}%
\hspace{0.05\textwidth}
\begin{subfigure}{.45\textwidth}
    \centering
    \includegraphics[width=0.95\linewidth]{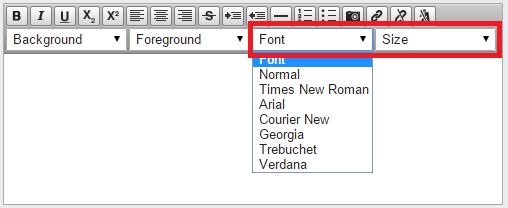}
    \caption{Q: I have been trying to add custom fonts and custom sizes in select boxes provided with the 'RichTextArea' of Vaadin. <IMAGE> How do I do this?}
    \label{fig:example_b}
\end{subfigure}
\caption{Examples of a required image (left) and a related image that is not required to answer the question (right).}
\label{fig:example}
\end{figure*}

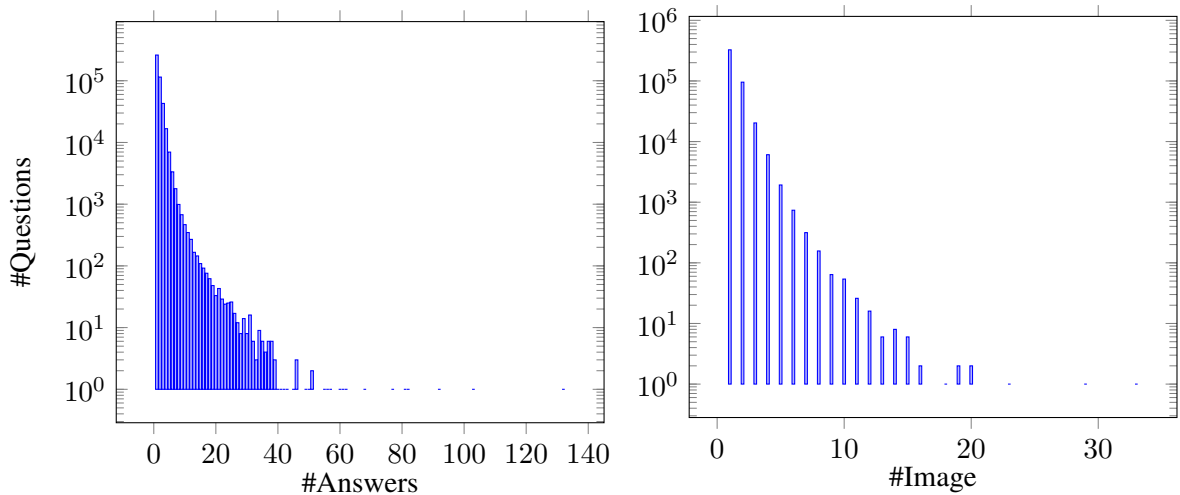
\begin{figure*}[t]
\begin{tikzpicture}
\begin{axis}[
    width=0.5\textwidth,
    bar width=1,
    ybar,
    ylabel={\#Questions},
    ymode=log,
    xlabel={\#Answers},
]
    \addplot coordinates {(1, 261099) (2, 114766) (3, 42926) (4, 16726) (5, 6984) (6, 3342) (7, 1783) (8, 990) (9, 675) (10, 466) (11, 347) (12, 269) (13, 166) (14, 145) (15, 109) (16, 92) (17, 76) (18, 62) (19, 48) (20, 33) (21, 43) (22, 29) (23, 24) (24, 25) (25, 26) (26, 17) (27, 12) (28, 8) (29, 14) (30, 8) (31, 16) (32, 6) (33, 3) (34, 9) (35, 6) (36, 4) (37, 6) (38, 6) (39, 3) (40, 1) (41, 1) (42, 1) (43, 1) (44, 0) (45, 1) (46, 3) (47, 0) (48, 0) (49, 1) (50, 1) (51, 2) (52, 0) (53, 0) (54, 0) (55, 1) (56, 1) (57, 1) (58, 0) (59, 0) (60, 1) (61, 1) (62, 1) (63, 0) (64, 0) (65, 0) (66, 0) (67, 0) (68, 1) (69, 0) (70, 0) (71, 0) (72, 0) (73, 0) (74, 0) (75, 0) (76, 0) (77, 1) (78, 0) (79, 0) (80, 0) (81, 1) (82, 1) (83, 0) (84, 0) (85, 0) (86, 0) (87, 0) (88, 0) (89, 0) (90, 0) (91, 0) (92, 1) (93, 0) (94, 0) (95, 0) (96, 0) (97, 0) (98, 0) (99, 0) (100, 0) (101, 0) (102, 0) (103, 1) (104, 0) (105, 0) (106, 0) (107, 0) (108, 0) (109, 0) (110, 0) (111, 0) (112, 0) (113, 0) (114, 0) (115, 0) (116, 0) (117, 0) (118, 0) (119, 0) (120, 0) (121, 0) (122, 0) (123, 0) (124, 0) (125, 0) (126, 0) (127, 0) (128, 0) (129, 0) (130, 0) (131, 0) (132, 1)};
\end{axis}
\end{tikzpicture}
\begin{tikzpicture}
\begin{axis}[
    width=0.5\textwidth,
    bar width=1,
    ybar,
    ymode=log,
    xlabel={\#Image},
]
    \addplot coordinates {(1, 326154) (2, 95534) (3, 20303) (4, 6078) (5, 1925) (6, 738) (7, 315) (8, 157) (9, 64) (10, 54) (11, 26) (12, 16) (13, 6) (14, 8) (15, 6) (16, 2) (17, 0) (18, 1) (19, 2) (20, 2) (21, 0) (22, 0) (23, 1) (24, 0) (25, 0) (26, 0) (27, 0) (28, 0) (29, 1) (30, 0) (31, 0) (32, 0) (33, 1)};
\end{axis}
\end{tikzpicture}
\centering
\caption{Frequency of questions based on answer counts (left) and image counts (right) in StackOverflowVQA dataset}
\label{fig:histograms}
\end{figure*}

We develop StackOverflowVQA by processing and filtering the \citet{Mike_2023} dataset which is available on Hugging Face. The dataset has 58.3M rows including questions, answers, and some wiki-related posts submitted to StackOverflow before June 14th of 2023 formatted as Markdown text. The data is sourced from \citet{Stack_Exchange_Community_2023}. The total number of questions in this dataset is 23,536,500.\\
This dataset can be challenging because the model has to have programming knowledge as well as an understanding of the image contents. Also, the questions and the answers can be very long (See \#WpQ and \#WpA in Table \ref{tab:statistics}).\\

\subsection{StackOverflowVQA}
StackOverflowVQA includes all questions that have at least one image as well as their answers (at least one answer) which has 451,394 questions with their corresponding answers (799,270 answers in total). Figure \ref{fig:histograms} shows the frequency of questions based on answer counts and image counts. As can be seen, there are lots of questions with multiple answers and images. The presence of multiple human-generated answers per question is a distinctive feature of our dataset.

To understand whether images are essential to answering questions, we select 100 random samples from StackOverflowVQA and manually examine whether the accompanying image is required to answer the question. One percent of the images are memes or similar unrelated images while 63 percent of the samples have images that are required to answer the question correctly. While the remaining 36 percent are related images but the questions can be answered correctly without them (Figure \ref{fig:qc} shows the distribution and figure \ref{fig:example} shows examples of an image that is required for answering the question and one that is not required).

\subsection{StackOverflowVQA-filtered}

This version of the dataset only includes the questions which have only one image and have an accepted answer. Also, we download the images, and if the image of a question can not be downloaded, we omit that question as well. This dataset contains 204,574 question-answer pairs which have been split to train and test using 90\% and 10\% of the samples respectively.

\subsection{StackOverflowVQA-filtered-small}

Due to the unavailability of resources, we use a subset of the filtered version of the dataset to fine-tune the GIT model. This subset has only 10\% of the filtered dataset and the question + answer tokens of it are not more than 512. This dataset has 18,412 and 2,046 question-answer pairs for training and testing respectively.

\section{Model}

We use GIT as the baseline model. This model is trained on general images and is not ideal for images in our dataset because they are mostly computer screenshots. Also, the texts are programming questions and answers while the model is trained on general text. So, this dataset is out of domain data for the model.\\
Furthermore, It is also not suitable for the texts because it has a limit of 512 tokens but with our StackOverflowVQA-filtered-small dataset, this problem can be solved.\\
The model gets the image and the question as the input and generates the answer using them.

\section{Training}

We first split 10\% of the training set as the validation set and use it to find the optimal
number of epochs. Then, we use the whole training set to fine-tune the model and evaluate it using the test set. We use a gradient clip with the value of 12, an AdamW optimizer with default hyper-parameters and a learning rate of 5e-5, a cosine schedule without warm-up steps, and a batch size of 8 (due to limited resources).

\section{Results}

Table \ref{tab:results} shows the preliminary results. The results show that the introduced dataset is challenging for even a state-of-the-art model like GIT. The low scores for more n-grams with $n>1$ show that the model does not have any programming knowledge to produce a coherent answer.\\
Also, these results can show the inefficiency of the evaluation metrics for such tasks and the need to design a better evaluation metric.\\
For sure, training the model with the whole dataset can improve the results.

\section{Future Work}

The purpose of this work is to attract researchers to a new application of visual question answering. One can do the following suggested actions to improve the results.

\begin{itemize}
    \item The baseline accepts a fixed number of images and because of that we selected the samples which have only one image. This limitation reduces the accuracy of the model.
    \item The baseline has been trained on general images and text but the images of the proposed dataset are mostly screenshots and texts are programming questions and answers. A pretraining on such image-text pairs can significantly improve the results.
    \item Most of the StackOverflow questions have more than one answer which in most cases are relevant and can help the model to gain programming knowledge.
    \item Also, the text-only question-answer pairs can teach the model programming.
    \item The position of the images in the question can be important especially when we have more than one image.
    \item Some of the answers have images as well, future work can work on generating the images in answer.
\end{itemize}

\section{Conclusion}

In this work, we introduce a new dataset called StackOveflowVQA which consists of StackOverflow questions that have at least one image in their body. This can be used to train a model to answer the programming questions and use screenshots or other related images to help the model improve its accuracy. Also, we use the GIT model as a baseline on a small subset of this dataset.


\end{document}